%% file: acl2023.tex
\pdfoutput=1

\documentclass[11pt]{article}

\usepackage[]{ACL2023}

\usepackage{times}
\usepackage{latexsym}

\usepackage{times}
\usepackage{siunitx}
\usepackage{amssymb}

\usepackage{todonotes}
\newcommand*{\yoruba}{Yor\`ub\'a\xspace}

\usepackage{microtype}

\usepackage{pgfplots}
\usepackage{tipa}
\usepackage{tikz}
\usepackage{rotating}
\usepackage{xspace}

\usepackage{makecell}

\definecolor{bblue}{HTML}{4F81BD}
\definecolor{rred}{HTML}{C0504D}
\definecolor{ggreen}{HTML}{9BBB59}
\usepackage[T1]{fontenc}

\usepackage[utf8]{inputenc}

\usepackage{microtype}

\usepackage{inconsolata}

\usepackage{booktabs}
\usepackage{graphicx}
\usepackage[caption=false]{subfig}
\usepackage{cleveref}
\usepackage{multirow}
\usepackage{amssymb}
\usepackage{pifont}
\newcommand{\cmark}{\ding{51}}%
\newcommand{\xmark}{\ding{55}}%

\usepackage{pgfplots}
\usepackage{tikz}

\usepackage{color, colortbl}
\newcommand{\best}{\cellcolor{blue!30}}

\usepackage{xspace}

\definecolor{Gray}{gray}{0.9}

\title{SemEval-2023 Task 12: Sentiment Analysis for African Languages (AfriSenti-SemEval)}

\author{Shamsuddeen Hassan Muhammad$^{1,2+*}$, Idris Abdulmumin$^{3+*}$, Seid Muhie Yimam$^{4*}$,\\
\bf David Ifeoluwa Adelani$^{5*}$, Ibrahim Sa'id Ahmad$^{2+*}$, Nedjma Ousidhoum$^{6}$, \\
\bf Abinew Ayele$^{4,7}$, Saif M. Mohammad$^{8}$, Meriem Beloucif$^{9}$
, Sebastian Ruder$^{10}$
\\
    \footnotesize $^1$University of Porto, $^2$Bayero University Kano, $^3$Ahmadu Bello University, Zaria, $^4$Universität Hamburg, \\
    \footnotesize $^5$University College London, $^6$University of Cambridge, $^7$Bahir Dar University, $^{8}$National Research Council Canada, \\
    \footnotesize $^9$Uppsala University, $^{10}$Google Research, $^{*}$Masakhane NLP, $^{+}$HausaNLP\\
    \footnotesize \texttt{shmuhammad.csc@buk.edu.ng}
}

\begin{document}
\maketitle 
\begin{abstract}
We present the first Africentric SemEval Shared task, Sentiment Analysis for African Languages (AfriSenti-SemEval)\footnote{The dataset is available at \url{https://github.com/afrisenti-semeval/afrisent-semeval-2023}.}. 
AfriSenti-SemEval is a sentiment classification challenge in 14 
African languages~(Amharic, Algerian Arabic, Hausa, Igbo, Kinyarwanda, Moroccan Arabic, Mozambican Portuguese, Nigerian Pidgin, Oromo, Swahili, Tigrinya, Twi, Xitsonga, and \yoruba) \cite{muhammad2023afrisenti}, using data labeled with 3 sentiment classes.
We present three subtasks: (1)\ Task A: monolingual classification, which received 44 submissions; (2)\ Task B: multilingual classification, which received 32 submissions; and (3) Task C: zero-shot classification, which received 34 submissions. 
 The best performance for tasks A and B was achieved by NLNDE team with 71.31 and 75.06 weighted F1, respectively. 
UCAS-IIE-NLP achieved the best average score for task C with 58.15 weighted F1. 
    We describe the various approaches adopted by the top 10 systems and their approaches.

\end{abstract}

\section{Introduction}

Sentiment Analysis is a prominent sub-field of Natural Language Processing that focuses on the automatic identification of sentiments or opinions expressed through online content, such as social media posts, blogs, or reviews \cite{liu2020sentiment,SentimentEmotionSurvey2021, nakov-etal-2016-semeval}. 
Example applications are the computational analysis of emotions in language, which has been applied to literary analysis and culturonomics \cite{mohammad-2011-upon,emotionarcs,hamilton2016diachronic}; commercial use (e.g., tracking opinions towards products); and research in psychology and social science \cite{dodds2015human,mohammad-etal-2016-semeval}.
Despite tremendous amount of  work in sentiment analysis over the last two decades, little work has been conducted on under-represented languages in general and African languages in particular.

\begin{figure}
    \centering
    \includegraphics[scale=0.65]{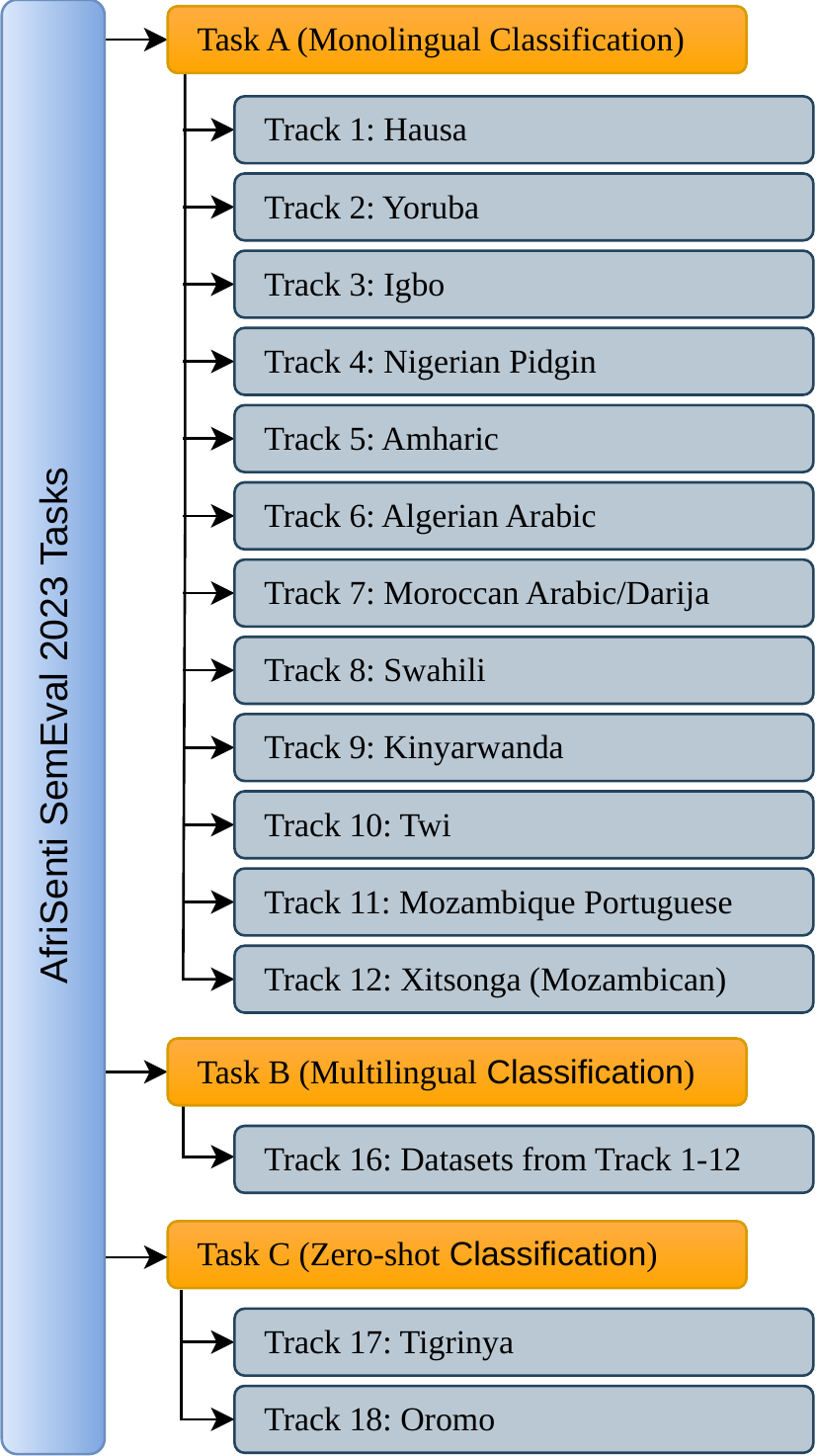}

    \caption{The AfriSenti-SemEval 2023 Shared Task tracks.}
    \label{fig:semeval_tasks}
\end{figure}
%
%

\nocite{nakov-etal-2013-semeval,pontiki-etal-2014-semeval,ghosh-etal-2015-semeval,nakov-etal-2016-semeval,semeval-2017-international,mohammad-etal-2018-semeval,chatterjee-etal-2019-semeval,patwa-etal-2020-semeval,meaney-etal-2021-semeval,barnes-etal-2022-semeval}
\begin{figure*}
    \centering
    \includegraphics[width=0.9\textwidth]{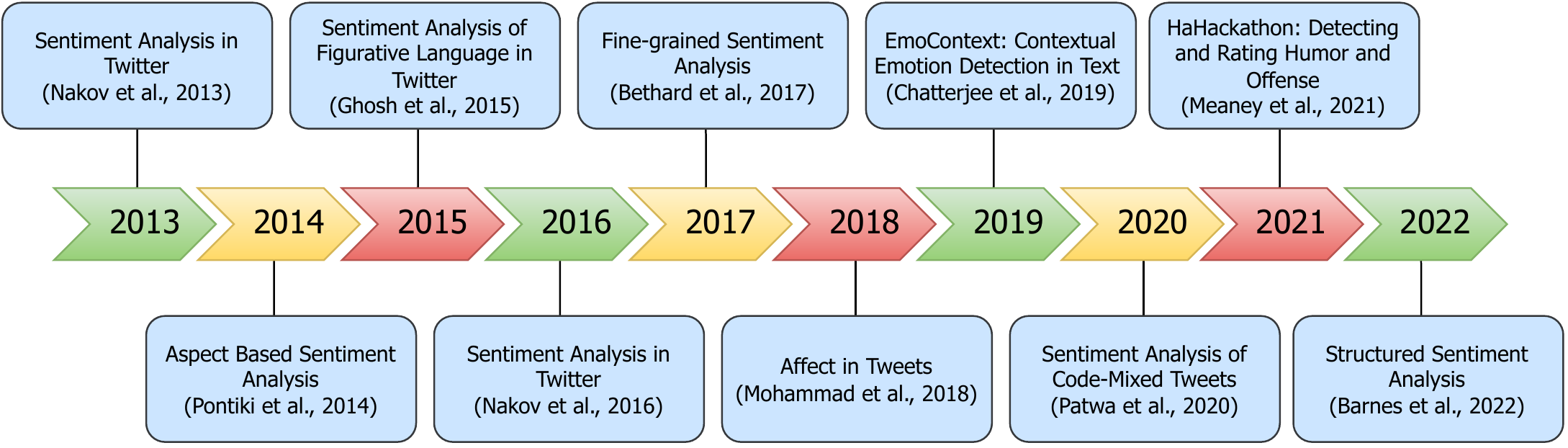}
     \caption{A timeline of SemEval Shared Tasks from 2013 to 2022 with examples of sentiment analysis tasks.}
    \label{fig:semeval_timeline}
\end{figure*}

Africa has a long and rich linguistic history, experiencing language contact, language expansion, development of trade languages, language shift, and language death, on several occasions. The continent is incredibly linguistically diverse and home to over 2000 languages. This includes 75 languages with at least one million speakers each. Africa has a rich tradition of storytelling, poems,
songs, and literature \citep{storytellingAfrica, africaStoryTelling}.
Yet, it is only in recent years that there is nascent interest in NLP research for African languages, including Named Entity Recognition \cite[NER;][]{adelani-etal-2021-masakhaner,adelani2022masakhaner, 10040676}, Machine Translation \cite[MT;][]{nekoto-etal-2020-participatory,abdulmumin-EtAl:2022:WMT,adelani2022findings,9971385},  and Language Identification (LID) for African languages~\cite{adebara2022afrolid}. However, African sentiment analysis 
has not yet received comparable attention. 
Similarly, although sentiment analysis is a common task in SemEval (see tasks examples in \Cref{fig:semeval_timeline}), 
previous tasks have mainly focused on high-resource languages.

To this end, we present the \textit{AfriSenti-SemEval}, a shared task in the 2023 edition of the Semantic Evaluation workshop \cite{SemEval:2023}. \textit{AfriSenti-SemEval} targets sentiment analysis in 
low-resource African languages. We provide researchers interested in African NLP with 110K sentiment-labeled tweets that were collected using the Twitter API.
These tweets are in 14 languages (Amharic, Algerian Arabic, Hausa, Igbo, Kinyarwanda, Moroccan Arabic, Mozambican Portuguese, Nigerian Pidgin, Oromo, Swahili, Tigrinya, Twi, Xitsonga, and Yoruba) from four language families (Afro-Asiatic, English Creole, Indo-European and Niger-Congo). The annotations were conducted by native speakers of the respective languages. Besides making the annotated dataset public, we also share 
sentiment lexicons in most of the languages.

%

AfriSenti-SemEval 2023 consists of 15 tracks from three sub-tasks on the 14 collected datasets as illustrated in \Cref{fig:semeval_tasks}. We  received 
submissions from 44 teams, 
with 29 submitting a system description paper. 
The top-ranked teams for the different subtasks used pre-trained language models (PLMs). In particular, AfroXLMR \cite{alabi-etal-2022-adapting}, an Africa-centric model, was the best performing model in both Tasks A (monolingual) and B (multilingual) with an average weighted F1 score of 71.30\% and 75.06\% respectively. For Task C, sentiment lexicons were used to build a lexicon-based multilingual BERT, which performed best in this setting with an average weighted F1 of 58.15\%. 

\section{ Background and Related Tasks}


Early work in sentiment analysis relied on lexicon-based  approaches \cite{Turney2002ThumbsUO, Taboada2011LexiconBasedMF, mohammad-etal-2013-nrc}. Subsequent work employed more advanced machine learning \cite{agarwal2016machine,Le2020TwitterSA}, deep learning \cite{zhang2018deep,yadav2020sentiment}, and hybrid approaches that combine lexicon and machine learning-based approaches \cite{gupta2020enhanced, kaur2022incorporating}. Nowadays, pre-trained language models (PLMs), such as XLM-R~\cite{conneau-etal-2020-unsupervised}, mDeBERTaV3~\cite{he2021debertav3}, AfriBERTa~\cite{Ogueji2021SmallDN}, AfroXLMR ~\cite{alabi-etal-2022-adapting} and XLM-T~\cite{barbieri-etal-2022-xlm} provide state-of-the-art performance for sentiment classification in different languages. 



Recent work in sentiment analysis focused on sub-tasks that tackle new challenges, including aspect-based \cite{chen2022discrete}, 
multimodal \cite{liang2022msctd}, 
explainable \cite{cambria-etal-2022-senticnet}, and multilingual sentiment analysis \cite{muhammad-etal-2022-naijasenti}. 
On the other hand, standard sentiment analysis sub-tasks such as polarity classification (positive, negative, neutral) are widely considered saturated and almost solved \cite{Poria2020BeneathTT}, with an accuracy of 97.5\% in certain domains \cite{raffel2020exploring,jiang-etal-2020-smart}. However, while this may be true for high-resource languages in relatively clean, long-form text domains such as movie reviews, noisy user-generated data in low-resource languages still presents a challenge \cite{Yimam2020ExploringAS}. Additionally, African languages exhibit new challenges for sentiment analysis such as dealing with tone, code-switching, and digraphia \cite{adebara-abdul-mageed-2022-towards}.
Thus, further research is necessary to assess the efficacy of existing 
NLP techniques and present solutions that can solve language-specific challenges in African contexts. SemEval, with its widespread recognition and popularity, is an ideal venue to conduct a shared task in sentiment analysis in African languages. 

The SemEval competition has become the de facto venue for sentiment analysis shared tasks, featuring at least one task per year as shown in \Cref{fig:semeval_timeline}. Some tasks focused on three-way sentiment classification---positive, negative, or neutral---while others explored more fine-grained aspect-based sentiment analysis \cite[ABSA;][]{rosenthal_semeval-2014_nodate, rosenthal_semeval-2015_2019,nakov-etal-2016-semeval,patwa-etal-2020-semeval}. Additionally, there are other closely related tasks, including the \textit{Affect in Tweets} task, which involves inferring perceived emotional states of a person from their tweet \cite{mohammad-etal-2018-semeval} and stance detection \cite{mohammad-etal-2016-semeval}, which refers to the automatic identification of stance of an author towards a target from text, where the stance can be in favor, against, or neutral. Finally, structured sentiment analysis requires participants to predict the sentiment graphs present in a text. Each sentiment graph comprises a sentiment holder, a target, an expression, and a polarity \cite{barnes-etal-2022-semeval}. 

\section{Task Description and Settings}

The AfriSenti-SemEval shared task consists of three sub-tasks: A)\ monolingual sentiment classification, B)\ multilingual sentiment classification, and C)\ zero-shot sentiment classification. 
As shown in \Cref{fig:semeval_tasks}, each sub-task also includes one or more tracks depending on the languages involved. 
Participants were free to participate in one or more sub-tasks and one or more tracks for each chosen subtask.\\[-20pt] 
\begin{description}
    \item[Task A: Monolingual Sentiment Classification] 
    Given a training set in a language, determine the polarity (positive/negative/neutral) of tweets in the same language. If a tweet conveys both positive and negative sentiments, the strongest sentiment should be chosen. This sub-task involves 12 tracks (all languages except Oromo and Tigrinya), with one track per language.\\[-20pt]
    

    \item[Task B: Multilingual Sentiment Classification] Given the combined training sets from Task A, determine the polarity (positive/negative/neutral) of tweets in the test sets of the languages. This sub\-task has only one track with tweets from 12 languages (Hausa, Yoruba, Igbo, Nigerian\_Pidgin, Amharic, Algerian Arabic, Moroccan Arabic Darija, Swahili, Kinyarwanda, Twi, Mozambican Portuguese, and Xitsonga).

    \item[Task C: Zero-Shot Sentiment Classification] Given unlabelled tweets in two African languages (Tigrinya and Oromo), use any of the training datasets of Task A to determine the sentiment of a tweet in the two target languages. This sub-task has two tracks (Tigrinya and Oromo).

\end{description}


\subsection{Pilot Dataset}

We released the pilot datasets for our SemEval shared task one month before the start of the shared task. The pilot datasets allowed the participants to have a better understanding of the shared task (i.e.,\ the datasets, the languages involved, and the labels).

\subsection{Task Settings}
The AfriSenti-SemEval shared task consisted of two phases: (1)\ the development phase and (2)\ the evaluation phase. In the development phase, we released a training set with gold labels and a development set without gold labels. Participants trained their models on the training set, tested on the development set, and submitted their predictions on the \href{https://codalab.lisn.upsaclay.fr/competitions/7320}{CodaLab} competition page for evaluation. The task offers a prize to the best-performing team in each of the three sub-tasks (A, B, and C) based on the following criteria\footnote{\url{https://github.com/afrisenti-semeval/afrisent-semeval-2023}}: (1) African League---for teams with at least one African member to encourage African participation; (2) Students League---for Master's and Undergraduate students only; and (3) Worldwide League---open to all participants.



\section{Dataset and Lexicon}


\begin{table*}[!htb]
    \centering
    \resizebox{\textwidth}{!}{
        \small
        \begin{tabular}{@{}lllrrrc@{}}
        \toprule[1.5pt]
        
        \multirow{2}{*}{\textbf{Language}} & \multirow{2}{*}{\textbf{Subregion}} & \multirow{2}{*}{\textbf{Script}} & \multicolumn{3}{c}{\textbf{Dataset}} & \multirow{2}{*}{\makecell{\textbf{Manual}\\\textbf{Lexicon}}}
        \\ \cmidrule{4-6}
        & & & \textbf{Train} & \textbf{Dev} & \textbf{Test} & \\ \midrule
        Amharic\texttt{(amh)} & East Africa & Ethiopic & 5,985 & 1,498 & 2,000 & \cmark \\
        Algerian Arabic/Darja \texttt{(arq)} & North Africa & Arabic &  1,652 & 415 & 959 & \xmark \\
         Hausa \texttt{(hau)} & West Africa  & Latin &  14,173 & 2,678 & 5,304  
         & \cmark \\
        Igbo \texttt{(ibo)} & West Africa   & Latin  &  10,193 & 1,842 & 3,683 
        & \cmark \\
        Kinyarwanda \texttt{(kin)} & East Africa  & Latin & 3,303 & 828 & 1,027    
        & \xmark \\
        Moroccan Arabic/Darija \texttt{(ary)} & North Africa  & Arabic/Latin & 5,584 & 1,216 & 2,962   
        & \xmark \\
        Mozambican Portuguese \texttt{(pt-MZ)} & Southeast Africa  & Latin  & 3,064 & 768 & 3,663  
        & \xmark\\
        Nigerian Pidgin \texttt{(pcm)} & West Africa  & Latin & 5,122 & 1,282 & 4,155  
        & \cmark \\
        Oromo \texttt{(orm)} & East Africa  & Latin & - & 397 & 2,096 
        & \cmark \\
        Swahili \texttt{(swa)} & East Africa  & Latin & 1,198 & 454 & 749 
        & \xmark \\ 
        Tigrinya \texttt{(tir)} & East Africa   & Ethiopic  & - & 399 & 2,001  
        & \cmark \\
        
        Twi  \texttt{(twi)} & West Africa & Latin & 3,482 & 389 & 950 & \xmark \\
        
        Xitsonga \texttt{(tso)} & South Africa  & Latin  & 805 & 204 & 255
        & \cmark \\
         \yoruba \texttt{(yor}) & West Africa  & Latin  & 8,523 & 2,091 & 4,516 
         & \cmark \\
        \bottomrule[1.5pt] 
        \end{tabular}
    }
    \caption{African languages included in our study \cite{Lewis2009EthnologueL,muhammad2023afrisenti}.}
    \label{tab:aflang}
    \vspace*{-2mm}
\end{table*}

\pgfplotstableread[row sep=\\,col sep=&]{
    language & positive & negative & neutral\\
    ama & 2103 & 3273 & 4104\\
    arq & 851 & 1590 & 582\\
    hau & 7329 & 7226 & 7597\\
    ibo & 4762 & 4013 & 6940\\
    kin & 1402 & 1788 & 1965\\
    ary & 3287 & 2874 & 3598\\
    pcm & 3652 & 6380 & 524\\
    pt-MZ & 1480 & 1633 & 4379\\
    swa & 908 & 319 & 1784\\
    tso & 601 & 446 & 214\\
    twi & 2277 & 1815 & 726\\
    yor & 6344 & 3296 & 5487\\
    orm & 616 & 850 & 1026\\
    tir & 704 & 1185 & 509\\
}\classDist

\begin{figure}
    \centering
    \resizebox{\columnwidth}{!}{%
        \begin{tikzpicture}
        \begin{axis}[ybar,
                width=25cm,
                height=15cm,
                ymin=0,
                ymax=8200,        
                ylabel={Number of tweets},
                ylabel near ticks,
                xlabel={Languages},
                xlabel near ticks,
                xtick=data,
                symbolic x coords={ama, arq, ary, hau, ibo, orm, pcm, pt-MZ, kin, swa, tir, tso, twi, yor},
                major x tick style = {opacity=0},
                minor x tick num = 1,
                minor tick length=2ex,
                legend entries={positive, negative, neutral},
                legend columns=3,
                legend style={draw=none,nodes={inner sep=3pt}},
                ylabel style={font=\fontsize{21}{24}\selectfont},
                xlabel style={font=\fontsize{21}{26}\selectfont},
                legend style={font=\fontsize{21}{26}\selectfont},
                yticklabel style = {font=\huge,xshift=0.5ex},
                xticklabel style = {font=\huge,yshift=0.5ex,rotate=45},
                nodes near coords,  
                every node near coord/.append style={font=\footnotesize, rotate=90, anchor=west},
            ]
        \addplot[draw=black,fill=bblue] table[x index=0,y index=1] \classDist;
        \addplot[draw=black,fill=rred] table[x index=0,y index=2] \classDist;
        \addplot[draw=black,fill=ggreen] table[x index=0,y index=3] \classDist;
        \end{axis}
        \end{tikzpicture}
    }
    \vspace*{-6mm}
    \caption{Label distributions for the AfriSenti datasets.}
    \label{fig:AfriSenti_class_dist}
    \vspace*{-4mm}
\end{figure}
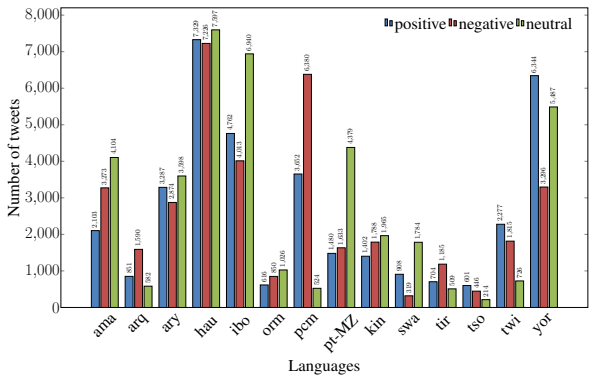

The AfriSenti collection covers 14 African languages, each with unique linguistic characteristics, writing systems, and language families, as shown in \Cref{tab:aflang}. The dataset covers 
 four of the five African sub-regions and includes the top three languages with the largest number of speakers in Africa (Swahili, Amharic, and Hausa). 
 The datasets include tweets collected using location-based and vocabulary-based (i.e., stopwords, sentiment lexicons, or language-specific terms) heuristics. \Cref{fig:AfriSenti_class_dist} shows the label distribution for the different languages in the AfriSenti dataset. 
 For more information on the AfriSenti dataset collection, annotation, and linguistic challenges, please refer to the AfriSenti dataset paper \cite{muhammad2023afrisenti}.
%

Early work on sentiment analysis in SemEval 
showed that sentiment lexicons can be leveraged and combined with training data and machine learning algorithms 
to obtain marked improvements in accuracy \cite{mohammad-etal-2013-nrc,NRCJAIR14}. Therefore, we also provide manually annotated sentiment lexicons in African languages\footnote{Dataset, scripts, lexicons available here:\\ \url{https://github.com/afrisenti-semeval/afrisent-semeval-2023}}. For languages that do not have manually curated lexicons, we translated existing lexicons into the target languages.
\Cref{tab:aflang} provides details on the sentiment lexicons in AfriSenti and indicates whether they were
manually created or translated.

\begin{table*}[]
    \centering
    \resizebox{\textwidth}{!}{
    \begin{tabular}{llrlrlr}
    \toprule
        & \multicolumn{2}{c}{\textbf{Sub-task A (monolingual)}} & \multicolumn{2}{c}{\textbf{Sub-task B (multilingual)}} & \multicolumn{2}{c}{\textbf{Sub-task C (zero-shot)}} \\ \cmidrule(lr){2-3}\cmidrule(lr){4-5}\cmidrule(lr){6-7}
        \textbf{\#} & \textbf{Team} & \textbf{F1} & \textbf{Team} & \textbf{F1} & \textbf{Team} & \textbf{F1} \\
    \midrule
    \rowcolor{Gray}
    * & AfriSenti baseline & 67.20 & AfriSenti baseline & 71.20  & AfriSenti baseline & 57.90 \\
    \midrule
        \textbf{1} & NLNDE & 71.31 & NLNDE & 75.06 & UCAS-IIE-NLP & 58.15 \\
        
        \textbf{2} & PALI & 70.28 & DN & 72.55 & NLNDE & 57.92 \\
        
        \textbf{3} & UM6P & 69.54 & UM6P & 71.95 & UM6P & 57.40 \\
        
        \textbf{4} & NLP-LISAC & 67.93 & GMNLP & 71.24 & DN & 55.19 \\
        
        \textbf{5} & GMNLP & 67.65 & Masakhane-Afrisenti & 70.34 & GMNLP & 51.70 \\
        
        \textbf{6} & ABCD Team & 67.51 & UCAS-IIE-NLP & 70.25 & ABCD Team & 51.59 \\
        
        \textbf{7} & UCAS-IIE-NLP  & 64.48 & Witcherses & 69.56 & Masakhane-Afrisenti &  50.04 \\
        
        \textbf{8} & HausaNLP   & 66.31 & HausaNLP & 69.54 & UBC-DLNLP & 49.41 \\
        
        \textbf{9} & UIO & 65.41 & ABCD Team & 69.22 & NLP-LISAC & 47.33 \\
        
        \textbf{10} & Masakhane-AfriSenti & 65.27 & NLP-LISAC & 67.30 & FIT BUT & 45.33 \\
        
    \bottomrule
    \end{tabular}
    }
    \caption{Top 10 submissions for Tasks A, B and C. We only ranked systems with  corresponding paper submissions. See \Cref{tab:affiliation} in \Cref{sec:aff} for paper information and teams' affliations.  \textbf{*} show the baseline result 
   from the AfriSenti dataset paper \cite{muhammad2023afrisenti}.}
    \label{tab:top_systems}
\end{table*}





\section{Evaluation}


All three tasks in the AfriSenti-SemEval shared task required participants to perform a sentiment (negative, neutral, positive) classification. To evaluate the performance of the systems submitted by the participating teams, we used weighted F1 as the evaluation metric.
For each label, weighted F1 measure its performance and weight it by the number of actual instances it has. This adjusts the ‘macro’ method to deal with uneven labels. We also provided the evaluation script to the participants to ensure consistency in the evaluation process.


We also created baseline systems for all three sub-tasks using multilingual pre-trained language models (PLMs).  
The baseline systems are: (1)\ monolingual baseline models based on multilingual PLMs for the 12 AfriSenti languages with training data; (2)\ models with multilingual training on all 12 languages, and evaluation on the combined test data of all 12 languages; and (3)\ zero-shot transfer of models to Oromo (\texttt{orm}) and Tigrinya (\texttt{tir}) from any of the 12 languages with available training data. Our best baseline model is shown in \Cref{tab:top_systems} and is based on \textit{fine-tuning} AfroXLMR-large\footnote{\url{https://huggingface.co/Davlan/afro-xlmr-large}}~\cite{alabi-etal-2022-adapting} in all the three sub-tasks. For more information on the baseline experimental results, please refer to the AfriSenti dataset paper \cite{muhammad2023afrisenti}.




\begin{table*}[t]
    \centering
    \footnotesize
    \resizebox{\textwidth}{!}{%
    \begin{tabular}{clrrrrrrrrrrrrr}
   \toprule[1.5pt]
         \# & Team & \textbf{\texttt{ama}} & \textbf{\texttt{arq}} & \textbf{\texttt{ary}} & \textbf{\texttt{hau}} & \textbf{\texttt{ibo}} & \textbf{\texttt{kin}} & \textbf{\texttt{pcm}} & \textbf{\texttt{pt-MZ}} & \textbf{\texttt{swa}} & \textbf{\texttt{tso}} & \textbf{\texttt{twi}} & \textbf{\texttt{yor}} & \textbf{Avg.}\\
    \midrule
        \textbf{1} & NLNDE  & 64.04 & 69.99 & \best\textbf{64.83} & \best\textbf{82.62} & \best\textbf{82.96} & \best\textbf{72.63} & 71.94 & 72.90 & \best\textbf{65.68} & \best\textbf{60.67} & 67.51 & 79.95 & \best\textbf{71.31} \\
        \textbf{2} & PALI & 65.56 & 72.62 & 55.92 & 81.10 & 81.30 & 69.61 & 75.16 & 73.83 & 64.37 & 56.26 & 67.58 & 80.06 & 70.28 \\
        \textbf{3} & king001 & 69.77 & 73.00 & 57.94 & 81.11 & 81.39 & 60.26 & 75.75 & 73.53 & 64.89 & 56.26 & \best\textbf{68.28} & \best\textbf{80.16} & 70.20 \\
        \textbf{4} & stce & 65.56 & 71.72 & 55.42 & 80.99 & 81.37 & 69.61 & 75.30 & 73.57 & 64.37 & 56.26 & 67.58 & 80.08 & 70.15 \\
        \textbf{5} & UM6P & 72.18 & 72.02 & 60.15 & 82.04 & 81.51 & 70.71 & 69.14 & 67.35 & 60.26 & 56.13 & 66.98 & 76.01 & 69.54 \\
        \textbf{6} & mitchelldehaven & 60.83 & 69.64 & 63.54 & 78.75 & 78.99 & 72.48 & 69.17 & 72.28 & 63.01 & 60.32 & 65.98 & 77.25 & 69.35 \\
        \textbf{7} & ymf924 & 69.83 & 67.83 & 57.12 & 75.88 & 72.49 & 70.88 & 72.48 & 73.80 & 64.23 & 55.38 & 62.56 & 75.94 & 68.20 \\
        \textbf{8} & NLP-LISAC & 64.71 & \best\textbf{74.20} & 62.11 & 79.74 & 79.66 & 65.59 & 68.00 & 65.90 & 59.69 & 53.72 & 66.38 & 75.42 & 67.93 \\
        \textbf{9} & GMNLP & \best\textbf{78.42} & 67.99 & 55.23 & 79.56 & 75.34 & 71.80 & 68.84 & 71.90 & 63.70 & 51.67 & 56.46 & 70.84 & 67.65 \\
        \textbf{10} & ABCD Team & 58.05 & 63.50 & 61.54 & 81.50 & 82.28 & 67.36 & 66.30 & 67.21 & 63.10 & 53.92 & 65.61 & 79.73 & 67.51 \\
        \textbf{11} & tmn & 63.11 & 70.54 & 57.22 & 72.93 & 72.66 & 69.82 & 69.63 & 73.09 & 61.66 & 53.66 & 67.06 & 75.08 & 67.21 \\
        \midrule
        \rowcolor{Gray}
        * & AfriSenti paper baseline & 61.60 & 68.30 & 56.60 & 80.70 & 79.50 & 70.60 & 68.70 & 71.60 & 63.40 & 47.30 & 64.30 & 74.10 & 67.20 \\
        \midrule
        \textbf{12} & PA14 & 64.52 & 72.01 & 55.34 & 77.17 & 80.29 & 61.16 & 75.53 & \best\textbf{74.98} & 61.73 & 50.46 & 67.39 & 65.65 & 67.19 \\
        \textbf{13} & UCAS-IIE-NLP & 67.45 & 63.31 & 55.36 & 80.79 & 78.51 & 71.47 & 66.63 & 65.13 & 60.73 & 51.09 & 60.15 & 77.13 & 66.48 \\
        \textbf{14} & HausaNLP & 57.30 & 65.12 & 58.49 & 80.97 & 76.96 & 70.61 & 68.48 & 68.37 & 63.23 & 50.27 & 64.07 & 71.88 & 66.31 \\
        \textbf{15} & UIO & 56.86 & 69.21 & 57.46 & 74.53 & 77.58 & 61.66 & 67.08 & 71.64 & 56.69 & 60.07 & 65.02 & 66.99 & 65.40 \\
        \textbf{16} & Masakhane-Afrisenti & 68.85 & 64.35 & 55.50 & 73.12 & 73.75 & 64.14 & 66.88 & 70.26 & 60.24 & 54.33 & 62.83 & 68.98 & 65.27 \\
        \textbf{17} & DN & 57.34 & 65.81 & 57.20 & 81.09 & 74.51 & 71.91 & 64.89 & 69.09 & 62.51 & 46.62 & 55.53 & 72.07 & 64.88 \\
        \textbf{18} & UMUTeam & 55.39 & 68.52 & 54.75 & 73.92 & 76.78 & 65.03 & 65.55 & 71.14 & 60.52 & 54.89 & 63.01 & 66.10 & 64.63 \\
        \textbf{19} & Howard University CS & 61.68 & 62.49 & 47.34 & 77.68 & 78.02 & 66.50 & 66.12 & 67.72 & 52.31 & 48.80 & 62.84 & 74.72 & 63.85 \\
        \textbf{20} & FIT BUT & 65.10 & 62.15 & 52.08 & 72.56 & 75.64 & 67.25 & 65.00 & 63.54 & 57.90 & 52.86 & 63.41 & 67.93 & 63.79 \\
        \textbf{21} & UBC-DLNLP & 56.88 & 64.02 & 53.06 & 79.37 & 77.52 & 62.02 & 65.57 & 61.98 & 58.60 & 45.49 & 65.14 & 71.02 & 63.39 \\
        \textbf{22} & Witcherses & 39.09 & 57.55 & 50.68 & 79.65 & 80.87 & 62.69 & 64.44 & 65.03 & 58.91 & 52.82 & 66.47 & 78.44 & 63.05 \\
        \textbf{23} & Sefamerve & 70.18 & 71.74 & 52.54 & 78.59 & 73.24 & 63.15 & 63.67 & 67.94 & 63.94 & 48.27 & 65.66 & 25.33 & 62.02 \\
        \textbf{24} & FUOYENLP & 54.25 & 68.14 & 52.78 & 73.17 & 72.22 & 54.20 & 68.01 & 72.88 & 54.31 & 51.38 & 57.94 & 64.31 & 61.97 \\
        \textbf{25} & efrat4050 & 31.99 & 55.40 & 54.43 & 75.65 & 76.28 & 59.56 & 53.17 & 62.75 & 51.76 & 54.10 & 64.18 & 72.76 & 59.34 \\
        \textbf{26} & JCT & 53.76 & 24.59 & 43.95 & 79.50 & 77.08 & 58.22 & 62.52 & 62.96 & 58.98 & 54.10 & 61.82 & 69.98 & 58.96 \\
        \textbf{27} & JacobLevy248 & 21.55 & 54.71 & 46.32 & 73.38 & 78.14 & 57.92 & 57.47 & 62.67 & 55.97 & 51.26 & 64.06 & 64.83 & 57.36 \\
        \textbf{28} & jacklight971 & 53.76 & 24.59 & 43.95 & 79.57 & 75.79 & 51.41 & 62.17 & 55.65 & 58.98 & 42.74 & 60.37 & 63.96 & 56.08 \\
        \textbf{29} & MaChAmp & 2.26 & 32.87 & 12.79 & 17.02 & 26.91 & 17.79 & 40.20 & 51.17 & 44.22 & 26.01 & 20.17 & 18.87 & 25.86 \\\hline
        \textbf{NR} & TechSSN & - & - & - & 80.32 & - & - & - & - & - & - & - & - & 80.32 \\
        \textbf{NR} & KINLP & - & - & - & - & - & 72.50 & - & - & - & - & - & - & 72.50 \\
        \textbf{NR} & afrisenti23kb & - & - & - & - & - & 71.00 & - & - & - & - & - & - & 71.00 \\
        \textbf{NR} & foul & - & 70.36 & - & - & - & - & - & - & - & - & - & - & 70.36 \\
        \textbf{NR} & Bhattacharya\_Lab & - & - & - & - & - & 47.10 & \best\textbf{75.96} & - & - & - & - & 79.86 & 67.64 \\
        \textbf{NR} & uid & - & - & 55.29 & 80.45 & 78.51 & 70.99 & - & - & - & 46.91 & 59.63 & 76.98 & 66.97 \\
        \textbf{NR} & TBS & - & - & 55.29 & 80.67 & 78.47 & 70.98 & - & - & - & 49.40 & 55.11 & 76.57 & 66.64 \\
        \textbf{NR} & DuluthNLP & - & - & - & - & - & - & 65.85 & - & - & - & 64.29 & - & 65.07 \\
        \textbf{NR} & Uppsala University & - & - & - & - & - & - & 64.69 & - & - & - & - & - & 64.69 \\
        \textbf{NR} & Seals\_Lab & 54.67 & - & - & 80.85 & 80.82 & - & - & 52.43 & 52.18 & - & - & - & 64.19 \\
        \textbf{NR} & NLP-LTU & - & - & 60.27 & - & - & - & - & - & - & - & - & - & 60.27 \\
        \textbf{NR} & Trinity & - & - & - & 76.53 & - & - & - & - & 47.55 & - & - & 50.00 & 58.03 \\
        \textbf{NR} & ronaharo & 14.98 & - & 39.72 & 72.84 & 73.21 & 57.87 & 62.52 & - & 53.35 & - & - & - & 53.50 \\
        \textbf{NR} & GunadarmaxBRIN & - & - & - & - & - & - & - & - & - & 50.11 & - & - & 50.11 \\
        \textbf{NR} & aptxaaaa & - & - & 51.01 & - & - & - & - & - & - & 48.97 & - & - & 49.99 \\
    \bottomrule
    \end{tabular}
    }
    \caption{Task A Results. The ranking is based on the average of the scores. Partial submissions were not included in the final ranking. (\textbf{NR} - No Ranking.)}
    \label{tab:task_a_results}
\end{table*}

\section{Participating Systems and Results}

The AfriSenti-SemEval competition had 213 registered participants on the CodaLab competition website. 
Of these, 44 teams submitted their systems during the evaluation phase. 
Out of the 44 submissions, 
29 submitted system-description papers. As participants could participate in one or more tasks, certain tasks received more submissions than others. Specifically, Task A (monolingual classification) had the highest number of participants with 44 submissions from different teams, followed by Task C (zero-shot classification) with 34 submissions and Task B (multilingual classification) with 33 submissions from different teams.

The majority of the teams participated in all tracks of each task, with 24 teams participating in at least 13 out of 15 tracks. For example, team NLNDE participated in all 12 tracks in Task A, one track in Task B, and two tracks in Task C. To rank the best-performing teams in each task and provide a comparison for future work, we rank each of the top-10 teams that participated in all tracks in a given task based on their average performance, as shown in \Cref{tab:top_systems}.

\Cref{tab:task_a_results}, \Cref{tab:task_b_results}, and \Cref{tab:task_c_results} present the overall results 
of participating systems for Task A (Monolingual), Task B (Multilingual), and Task C (Zero-shot), respectively. \Cref{tab:affiliation} in \Cref{sec:aff}
presents information regarding the teams and their affiliations. 
In the following sections, we describe the best systems in each subtask.

\subsection{Subtask A: Monolingual Sentiment Classification Systems}
We describe the top-10 teams that submitted system description papers as highlighted in \Cref{tab:top_systems}.

\paragraph{NLNDE \cite{wang-EtAl:2023:SemEval2}} used language adaptive pre-training (LAPT) and task adaptive pre-training (TAPT) as an additional pre-training step on AfroXLMR-large. The LAPT approach involved continued pre-training of the PLM on the monolingual portion of the Leipzig Corpus Collection~\citep{goldhahn-etal-2012-building} (covering Wikipedia, Community, Web, and News corpora) for the target language. TAPT involved continued pre-training on the AfriSenti training data of the target language. By leveraging LAPT followed by TAPT, they achieved 
significant improvements over fine-tuning AfroXLMR-large directly. \textbf{NLNDE ranked first in 7 out of 12 languages, and first in sub-task A}. 

\paragraph{PALI \cite{jin-EtAl:2023:SemEval}} used weighted fusion of several PLMs such as \textit{naija-twitter-sentiment-afriberta-large}\footnote{\url{https://huggingface.co/Davlan/naija-twitter-sentiment-afriberta-large}}, an AfriBERTa PLM trained on the NaijaSenti dataset~\cite{muhammad-etal-2022-naijasenti}, TwHIN-BERT~\cite{zhang2022twhin}, and mDeBERTaV3~\cite{he2021debertav3} trained on AfriSenti corpora translated to English. The performance of the PLMs varied across different languages. For example, AfriBERTa fine-tuned on NaijaSenti provided better results on Nigerian languages. 
 \textbf{PALI ranked first in 2 out of 12 languages, and second for sub-task A.} 

\paragraph{UM6P \cite{elmahdaouy-EtAl:2023:SemEval}} combined MARBERT~\cite{abdul-mageed-etal-2021-arbert} and an adapted AfroXLMR using a projection and a residual layer. First, AfroXLMR was first fine-tuned on the AfriSenti training data and web resources like MAFAND-MT~\cite{adelani-etal-2022-thousand} and WebCrawl African multilingual parallel corpora~\cite{vegi-etal-2022-webcrawl} using a whole-word masking objective~\cite{cui_whole_mask_bert}. It was then fine-tuned on the labeled AfriSenti training data and combined with MARBERT. The \textbf{UM6P team ranked 5th for sub-task A.}

\paragraph{NLP-LISAC \cite{benlahbib-boumhidi:2023:SemEval}} used different PLMs for different languages. For Hausa, Yoruba, Igbo, Swahili, and Kinyarwanda, they used mBERT~\cite{devlin-etal-2019-bert} that was adapted to African languages with continued pre-training. The checkpoints\footnote{\url{https://huggingface.co/models?search=davlan/xlm-roberta-base-finetuned}} were based on models released by~\citet{adelani-etal-2021-masakhaner}.  For Mozambican Portuguese, Xitsonga, Nigerian-Pidgin, and Amharic, they used mDeBERTaV3. They used DziriBERT~\citep{dziribert}, a model specifically trained for the Algerian dialect, for Algerian Arabic and AfriBERTa-large for Twi. \textbf{NLP-LISAC ranked first for Algerian Arabic}. 

\paragraph{GMNLP \cite{alam-EtAl:2023:SemEval}} used phylogeny-based adapter-tuning~\citep{faisal-anastasopoulos-2022-phylogeny} of AfroXLMR-large. They used a dictionary and machine translation-based data augmentation strategies to increase the amount of training data. The dictionary approach used PanLex~\cite{kamholz-etal-2014-panlex} while the MT approach was based on the WMT Shared task model~\cite{alam-anastasopoulos-2022-language}. However, this excluded some languages (Twi, Mozambican Portuguese, Nigerian Pidgin, Moroccan Arabic, and Algerian Arabic). Their findings suggest that using phylogeny adapters (language adapter, genus adapter, and family adapter) can lead to better performance. For example,  on Amharic, Hausa, Kinyarwanda, Nigerian-Pidgin, and Xitsonga, all the phylogeny adapters were useful. \textbf{GMNLP ranked first for Amharic}. 

\paragraph{ABCD Team \cite{dang-EtAl:2023:SemEval}} used soft voting ensemble of three PLMs: AfroXLMR, AfriBERTa, and LaBSE -- a multilingual sentence transformer model that supports over 100 languages, and is also popular for mining parallel corpus for machine translation. The \textbf{ABCD team ranked 10th in the sub-task A competition}. 

\paragraph{UCAS-IIE-NLP \cite{hu-EtAl:2023:SemEval}} used a lexicon-based multilingual transformer model based on AfroXLMR-base to facilitate language adaptation and sentiment-aware representation learning. Additionally, they applied a supervised adversarial contrastive learning strategy to improve the sentiment-spread representations and enhance model generalization. On average, their approach was worse than the AfriSenti baseline likely because they used AfroXLMR-base rather than the large version. Interestingly, they achieved much better results than the baseline on Amharic, Xitsonga, and \yoruba by over $3$ F1 points. 

\paragraph{HausaNLP \cite{abdullahi-EtAl:2023:SemEval}} used two BERT-based models: AfroXLMR-large and an Arabic BERT~\cite{inoue-etal-2021-interplay} fine-tuned on a sentiment corpus\footnote{\url{https://huggingface.co/CAMeL-Lab/bert-base-arabic-camelbert-da-sentiment}}. They used AfroXLMR-large for all languages except the Arabic dialects. On average across 12 languages, the HausaNLP system ranked lower than the AfriSenti paper baseline.  

\paragraph{UIO \cite{rnningstad:2023:SemEval}} compared four different PLMs: AfroXLMR-mini\footnote{\url{https://huggingface.co/Davlan/afro-xlmr-mini}} (based on language adaptation of MiniLMv2~\citep{wang-etal-2021-minilmv2}, a compressed XLM-R-large model), mpnet-base-v2 sentence transformer~\cite{reimers-gurevych-2019-sentence}, and XLM-T~\cite{barbieri-etal-2022-xlm} and XLM-T fine-tuned on a multilingual sentiment corpus\footnote{\url{https://huggingface.co/cardiffnlp/twitter-xlm-roberta-base-sentiment}}. They chose the latter as the best model since it was pre-trained both on the Twitter domain and a sentiment classification task. This aligns with the findings of \cite{muhammad2023afrisenti} where XLM-T was the second-best model because of in-domain pre-training. However, UIO's system underperformed the AfriSenti baseline as making use of a large Afro-centric PLM like AfroXLMR-large outperforms the use of domain and task-specific data with smaller models. 

\paragraph{Masakhane-AfriSenti \cite{azime-EtAl:2023:SemEval}} used five PLMs: LabSE, AfriBERTa, AfroXLMR-base, XLM-T, and Bernice~\cite{delucia-etal-2022-bernice}, a multilingual Twitter RoBERTa model. The performance of their best systems varied by language. For example, they reported AfroXLMR-base to be better for Hausa and Swahili, while LaBSE gave better results for Kinyarwanda and Mozambique Portuguese. In general, their best systems ranked lower than the AfriSenti baseline. 

Apart from the top-10 teams that submitted their papers, there were other teams that only worked on one or few languages, and achieved excellent rankings. For instance, \textbf{KINLP} \cite{nzeyimana:2023:SemEval} only attempted the task for the Kinyarwanda language. Their approach was based on KinyaBERT~\cite{nzeyimana-niyongabo-rubungo-2022-kinyabert}, a Kinyarwanda PLM that incorporates morphological features of the language during pre-training. \textbf{KINLP} ranked second for Kinyarwanda. \textbf{Bhattacharya\_Lab} \cite{hughes-EtAl:2023:SemEval} only worked on Nigerian-Pidgin and \yoruba. They pre-trained a RoBERTa-style~\cite{Liu2019RoBERTaAR} transformer-based architecture jointly on the two languages using the AfriBERTa training corpus and AfriSenti data. \textbf{Bhattacharya\_Lab ranked first for Nigerian-Pidgin, and fifth for \yoruba}.





\subsection{Subtask B: Multilingual Sentiment Classification Systems}
Most teams used the same model as in sub-task A for sub-task B with minor changes.
We highlight here the teams that 
used strategies apart from jointly training a PLM on the concatenation of all 12 sub-task A languages. We describe the top-10 teams with corresponding system description papers as shown in \Cref{tab:top_systems}. The complete results for this sub-task are shown in \Cref{tab:task_b_results}.

 \begin{table}[t]
    \centering
    \small
    \begin{tabular}{clr}
        \toprule
        \# & Team & F1 \\
        \midrule
        \textbf{1} & NLNDE & \best\textbf{75.06} \\
        \textbf{2} & king001 & 74.96 \\
        \textbf{3} & DN & 72.55 \\
        \textbf{4} & ymf924 & 72.34 \\
            \textbf{5} & mitchelldehaven & 72.33 \\
        \textbf{6} & UM6P & 71.95 \\
        \textbf{7} & GMNLP & 71.24 \\
        \midrule
        \rowcolor{Gray}
        \textbf{*} & AfriSenti baseline & 71.20 \\
        \midrule
        \textbf{8} & PA14 & 70.81 \\
        \textbf{9} & Masakhane-Afrisenti & 70.34 \\
        \textbf{10} & UCAS-IIE-NLP & 70.25 \\
        \textbf{11} & tmn & 70.05 \\
        \textbf{12} & Witcherses & 69.56 \\
        \textbf{13} & HausaNLP & 69.54 \\
        \textbf{14} & ABCD Team & 69.22 \\
        \textbf{15} & iREL & 68.23 \\
        \textbf{16} & BERT 4EVER & 67.46 \\
        \textbf{17} & NLP-LISAC & 67.30 \\
        \textbf{18} & Howard University CS & 66.60 \\
        \textbf{19} & ISCL\_WINTER & 66.38 \\
        \textbf{20} & UIO & 66.29 \\
        \textbf{21} & FIT BUT & 66.02 \\
        \textbf{22} & FUOYENLP & 65.51 \\
        \textbf{23} & UMUTeam & 65.47 \\
        \textbf{24} & GunadarmaxBRIN & 65.23 \\
        \textbf{25} & saroyehun & 64.34 \\
        \textbf{26} & JacobLevy248 & 64.04 \\
        \textbf{27} & efrat4050 & 63.85 \\
        \textbf{28} & Sefamerve & 63.40 \\
        \textbf{29} & DuluthNLP & 61.49 \\
        \textbf{30} & jacklight971 & 59.75 \\
        \textbf{31} & JCT & 59.50 \\
        \textbf{32} & Hercules & 59.35 \\
        \textbf{33} & ronaharo & 57.79 \\
        \bottomrule 
    \end{tabular}
    \caption{AfriSenti-SemEval Task B results.}
    \label{tab:task_b_results}
\end{table}

\paragraph{NLNDE} For each target language, they first chose the best source languages for multilingual training to prevent harmful interference from dissimilar languages. For selecting the source language set, they performed forward and backward source language selection, similar to feature selection approaches~\cite{pmlr-vR4-tsamardinos03a,Giorgos_fwd_bck_ioannis}. Forward feature selection starts with an empty set of languages and adds languages to it, while backward feature selection starts with a complete set of languages and then excludes languages from it. For example, the best source languages for multilingual training for Hausa using Forward selection are Kinyarwanda, Twi, Algerian Arabic and Nigerian-Pidgin. For \yoruba, the best source languages according to the backward selection are Kinyarwanda, Xitsonga, Twi, and Algerian-Arabic. NLNDE used multiple models for this task rather than a single one like most teams did. The target language of the tweet would determine the corresponding multiple training strategy. \textbf{NLNDE ranked first in this sub-task}. 

\paragraph{DN \cite{homskiy-maloyan:2023:SemEval}} They fine-tuned AfroXMLR-large on the 12 languages available in the training data. They performed additional pre-processing on the tweets before training, i.e., they removed links, hashtags, and @mentions, which boosted the performance of their system over those trained on a single multilingual model on all 12 languages. \textbf{DN ranked third in this sub-task}.


\paragraph{GMNLP}, unlike in sub-task A, did not use a phylogeny-based adapter fine-tuning for this sub-task due to the absence of language ID information. They only performed task adapter training. 



\paragraph{Witcherses \cite{gokani-srivatsa-mamidi:2023:SemEval}} used an ensemble of both multilingual PLMs and classical ML models: Logistic Regression, Random Forest, Support Vector Machine, AfriBERTa, AfroXLMR, and AfriBERTa fine-tuned on the NaijaSenti corpus. 

\paragraph{Hausa NLP} used the same approach as in sub-task A. They used AfroXLMR-large for multilingual training, which was previously fine-tuned on MasakhaNER 2.0\footnote{\url{https://huggingface.co/masakhane/afroxlmr-large-ner-masakhaner-1.0_2.0}} \citep{adelani2022masakhaner}. 


\paragraph{NLP-LISAC} used the same approach described in sub-task A. They chose the mDeBERTaV3 PLM to fine-tune on the multilingual corpus. 

\paragraph*{}
The other teams i.e. \textbf{UM6P},  \textbf{Masakhane-AfriSenti}, \textbf{UCAS-IIE-NLP}, and \textbf{ABCD Team} used the same approach as the one they used for sub-task A. The only difference was that they trained a PLM on all 12 languages instead of training a monolingual sentiment model. 


 \subsection{Subtask C: Zero-Shot Sentiment Classification Systems}

We provide an overview of the top-10 submissions with system description papers in \Cref{tab:top_systems} and show the complete results for this sub-task in \Cref{tab:task_c_results}.

\begin{table}[t]
    \centering
    \small
    \begin{tabular}{clrrr}
        \toprule
        \# & Team & \textbf{\texttt{orm}} & \textbf{\texttt{tir}} & \textbf{Avg.} \\
        \midrule
        \textbf{1} & UCAS-IIE-NLP & 45.82 & 70.47 & \best\textbf{58.15} \\
        \textbf{2} & NLNDE & 44.97 & \best\textbf{70.86} & 57.92 \\
        \midrule
        \rowcolor{Gray}
         * & AfriSenti baseline & \textbf{47.10} & 68.60 & 57.90 \\
         \midrule
        \textbf{3} & ymf924 & 45.34 & 70.39 & 57.87 \\
        \textbf{4} & UM6P & 45.27 & 69.53 & 57.40 \\
        \textbf{5} & TBS  & 45.12 & 69.61 & 57.37 \\
        \textbf{6} & uid  & 44.75 & 69.90 & 57.33 \\
        \textbf{7} & mitchelldehaven & \best 46.23 & 66.96 & 56.60 \\
        \textbf{8} & tmn & 41.97 & 69.09 & 55.53 \\
        \textbf{9} & DN & 41.45 & 68.93 & 55.19 \\
        \textbf{10} & GMNLP & 41.87 & 61.52 & 51.70 \\
        \textbf{11} & ABCD Team & 42.64 & 60.53 & 51.59 \\
        \textbf{12} & BERT 4EVER & 42.32 & 60.37 & 51.35 \\
        \textbf{13} & Masakhane-Afrisenti & 42.09 & 57.99 & 50.04 \\
        \textbf{14} & UBC-DLNLP & 41.79 & 57.03 & 49.41 \\
        \textbf{15} & NLP-LTU & 36.89 & 61.48 & 49.19 \\
        \textbf{16} & NLP-LISAC & 42.02 & 52.64 & 47.33 \\
        \textbf{17} & FIT BUT & 38.24 & 52.42 & 45.33 \\
        \textbf{18} & PA2023 & 27.76 & 62.27 & 45.02 \\
        \textbf{19} & saroyehun & 37.79 & 45.19 & 41.49 \\
        \textbf{20} & MaChAmp & 38.02 & 44.86 & 41.44 \\
        \textbf{21} & FUOYENLP & 35.57 & 47.18 & 41.38 \\
        \textbf{22} & Sefamerve & 39.35 & 34.16 & 36.76 \\
        \textbf{23} & PA14 & 39.86 & 29.49 & 34.68 \\\hline
        \textbf{NR} & yg.ren & - & 60.87 & 60.87 \\
        \textbf{NR} & hk001 & - & 60.35 & 60.35 \\
        \textbf{NR} & abc1234 & - & 59.07 & 59.07 \\
        \textbf{NR} & abc123 & - & 55.49 & 55.49 \\
        \textbf{NR} & abc1 & - & 42.42 & 42.42 \\
        \textbf{NR} & Bhattacharya\_Lab & 40.61 & - & 40.61 \\
        \textbf{NR} & Seals\_Lab & 40.61 & - & 40.61 \\
        \textbf{NR} & Snarci & 33.98 & - & 33.98 \\
        \textbf{NR} & abc12 & - & 33.33 & 33.33 \\
        \textbf{NR} & GunadarmaxBRIN & 32.05 & - & 32.05 \\
        \textbf{NR} & HausaNLP & 25.43 & - & 25.43 \\
        \bottomrule
    \end{tabular}
    \caption{Task C Results. The ranking is based on the average of the scores. Partial submissions are not included in the final ranking. (\textbf{NR} - No Ranking.)}
    \label{tab:task_c_results}
\end{table}

\paragraph{UCAS-IIE-NLP}  used the same approach described in sub-task A. They used additional lexicon information for zero-shot transfer to both Oromo and Tigrinya. \textbf{UCAS-IIE-NLP ranked first for sub-task C, first for Oromo, and second for Tigrinya.} Surprisingly, their best performance is below the AfriSenti baseline for Oromo ($-1.28$ F1), which is based on choosing the best language for zero-shot transfer. \citet{muhammad2023afrisenti} identified Hausa and Amharic as the best source languages for Oromo and Hausa and \yoruba as the best source languages for Tigrinya. Co-training on the two languages led to a better performance.

\paragraph{NLNDE} used the same approach as in sub-task B. They used forward and backward language selection  to decide the best source languages to transfer from. For Oromo, the best source languages they identified were Kinyarwanda, Hausa, \yoruba, and Xitsonga using forward selection, and Yoruba, Mozambique Portuguese, and Xitsonga using backward selection. Similarly, for Tigrinya, they identified Hausa, Kinyarwanda, Amharic, Moroccan Arabic, and Mozambique Portuguese in the forward selection, and Mozambican Portuguese, \yoruba, and Hausa. The languages selected were similar to those identified by \citet{muhammad2023afrisenti}. \textbf{NLNDE ranked second on sub-task C and first for Tigrinya}. 

\paragraph{Masakhane-AfriSenti}  used the multilingual model they introduced in sub-task B based on AfroXLMR-base and AfriBERTa. They also tried adapter-based training based on MAD-X~\cite{pfeiffer-etal-2020-mad}. Their final result is based on an ensemble of the three methods. 

\paragraph{UBC-DLNLP \cite{bhatia-EtAl:2023:SemEval}} compared about 10 PLMs for all the sub-tasks including XLM-R, AfriBERTa, mBERT, AfroXLMR, Serengeti~\citep{Adebara2022SERENGETIMM}, Serengeti\_ft (Serengeti + TAPT), and AfroXLMR-base\_ft (AfroXLMR-base + TAPT). Their best-reported result is based on AfroXLMR-base\_ft. Surprisingly, AfroXLMR-base\_ft and AfroXLMR-base, which were pre-trained on 17 African languages outperformed Serengeti\_ft  \& Serengeti, which were pre-trained on 500+ African languages. This indicates that training on a large number of languages may suffer from the curse of multilinguality~\cite{conneau-etal-2020-unsupervised} without appropriate scaling since it is difficult to learn a good representation for 500 languages with a small PLM of 270M parameters. 

\paragraph{FIT BUT \cite{aparovich-EtAl:2023:SemEval}} used AfroXLMR-small with additional adversarial training but they only achieved average performance. This is probably due to the use of a small PLM for training. 

\paragraph*{}
Other teams like \textbf{NLP-LISAC}, \textbf{UM6P}, \textbf{DN}, \textbf{GMNLP}, and \textbf{ABCD} adopted approaches similar to those of sub-task B: they trained on all multilingual datasets and performed zero-shot evaluation on Oromo and Tigrinya. 


 






\section{Discussion}
We summarize some of the approaches that led to the best results in different sub-tasks.

\paragraph{Sub-task A} All of the top-10 best-performing teams with systems description papers employed multilingual pre-trained models, especially Afro-centric models. For example, eight of the ten teams make use of AfroXLM---one of the best-performing PLM for African languages. AfroXLMR-large with additional pre-training often led to the best results while multilingual PLMs like mDeBERTaV3 and LaBSE led to competitive results. A few teams used other PLMs, specifically trained on Arabic variants such as DziriBERT. Some teams also reported significant language-specific improvements using further domain and task-specific pre-training. For instance, the NLNDE team, which ranked first, used both language and task adaptive pre-training. UIO and Masakhane-AfriSenti also demonstrated the benefit of domain adaptive pre-training. In addition, PALI and Masakhane-AfriSenti showed that using a PLM that has already been trained on sentiment classification can help. Interestingly, other teams using an ensemble of different fine-tuned PLMs tended to perform worse, which highlights that the quality of individual models is important. 

\paragraph{Sub-task B} Most teams used a single multilingual PLM and fine-tuned it on all languages. In fact, most of the best-ranking teams used AfroXLMR-large as it performed well on sub-task A. The best-performing team for this task, NLNDE, chose to select the most appropriate languages to co-train for each language before performing multilingual training, highlighting the importance of the choice of source languages.

\paragraph{Sub-task C} 
UCAS-IIE-NLP ranked first and used a lexicon-based multilingual BERT. 
This shows the usefulness of leveraging sentiment lexicons as side information in building language models. However, their best performance was below the AfriSenti paper baseline for Oromo ($-1.28$ F1). 
 
The top-performing teams in each subtask were not affiliated with African institutions. They developed the best models despite a lack of language expertise. This highlights both the generality of existing models and adaptation paradigms as well as the need for a more collaborative approach to building more effective and inclusive solutions for Africa-centric sentiment analysis.


\section{Conclusion}

We presented the \textit{SemEval-2023 Task 12: Sentiment Analysis for African Languages}, the first SemEval shared task that focuses on sentiment analysis for African languages. 
The task included monolingual classification (in Amharic, Algerian Arabic, Hausa, Igbo, Kinyarwanda, Moroccan Arabic, Mozambican Portuguese, Nigerian Pidgin, Swahili, Twi, Xitsonga, and \yoruba), multilingual classification, and zero-shot classification (in Oromo and Tigrinya).
We described the task settings, datasets, and baselines.

We discussed the main findings of the 44 participating teams who submitted their systems, based on their system description papers (i.e., 29 papers) as well as our observations and analysis of some common errors.  
Overall, the best ranking teams used pre-trained language models (PLMs), with Africa-centric models such as AfroXLMR performing the best in the Task A (monolingual) and Task B (multilingual) classification tasks, with an average weighted F1 of
71.3\%, and 75.06\%, respectively. For Task C (zero-shot), the top team used lexicon-based multilingual BERT and achieved an average weighted F1 of 58.15.\%.
These scores indicate that there is still room for improvement in polarity classification in low-resource settings.

By sharing our insights, we aim to encourage researchers to work on under-resourced and under-studied African languages and help them improve the performance of current sentiment analysis systems. In the future, we will extend our task to more languages by building additional datasets.



\section{Ethics Statement}

People often express sentiment in unique and interesting ways. Thus, there is large amounts of person--person variation. Therefore, any automatic method for sentiment analysis will achieve different results on data from different people, from different domains, etc. 
We do not recommend the use of automatic methods of sentiment analysis (based on individual instances of text) to make important decisions that can impact an individual.
Instead, it is often better to use automatic sentiment analysis to determine broad trends of sentiment
across large amounts of data.
Sentiment analysis, like many other AI technologies, can be used not just for beneficial purposes, but also to cause harm such as using it to identify and suppress dissent.
There are several such ethical considerations that should be accounted for when developing and deploying sentiment analysis systems. We refer to \citet{Mohammad22AER,Mohammad23ethicslex} for a comprehensive discussion of ethical considerations relevant to sentiment and emotion analysis.


\bibliographystyle{acl_natbib}
\bibliography{anthology,custom,afrisenti_system_papers,2023.semeval-1.0-fixed}

\appendix

\newpage
\section{Algorithms Used}
Algorithms used by the participants for data pre-processing and for building the classification systems are shown in \Cref{fig:algorithm_stats}.
\input{algorithm-stats}

\newpage
\section{Tools Used}
Tools used by the participants to implement their systems are shown in \Cref{fig:tools_used}.
\input{tools-used}

\section{Affiliation}
\label{sec:aff}
\Cref{tab:affiliation} shows the participating teams, the tasks they made submissions for, their system description paper, and their affiliations.
\input{affiliation}

\end{document}

%% file: algorithm-stats.tex
\pgfplotstableread[row sep=\\,col sep=&]{
    Algo & Freq\\
    XLM-R & 146\\
    AfriBerta & 141\\
    AfroLM & 102\\
    mBERT & 81\\
    SVM & 33\\
    AfriTeva & 15\\
    XGBoost & 14\\
    XLM-T  & 6\\
    RF & 4\\
    Afro-XLM-R & 45\\
    mDeBERTa & 116\\
    RNN & 1\\
    dziribert & 1\\
    LR & 1\\
    MNB & 1\\
    twhin-bert-bas & 1\\
    Bertweet & 1\\
    DarijaBERT & 1\\
    LightBGM & 1\\
    kNN & 1\\
    BERNICE & 1\\
    RoBERTA & 1\\
    MARBERT & 1\\
    KinyaBERT & 1\\
}\algorithmClassStat

\pgfplotstableread[row sep=\\,col sep=&]{
    Algo & Freq\\
    AfroLM & 117\\
    AfriBerta & 100\\
    XLM-R & 87\\
    mBERT & 61\\
    SVM & 45\\
    AfriTeva & 32\\
    XGBoost & 18\\
    RNN & 16\\
    AfroXLMR-large & 2\\
    LR & 2\\
    NLLB-200 & 2\\
    XLM-T & 2\\
    CNN & 1\\
    LSTM & 1\\
    Bertweet & 1\\
    KinyaBERT & 1\\
    RF & 1\\
}\algorithmPreProcStat

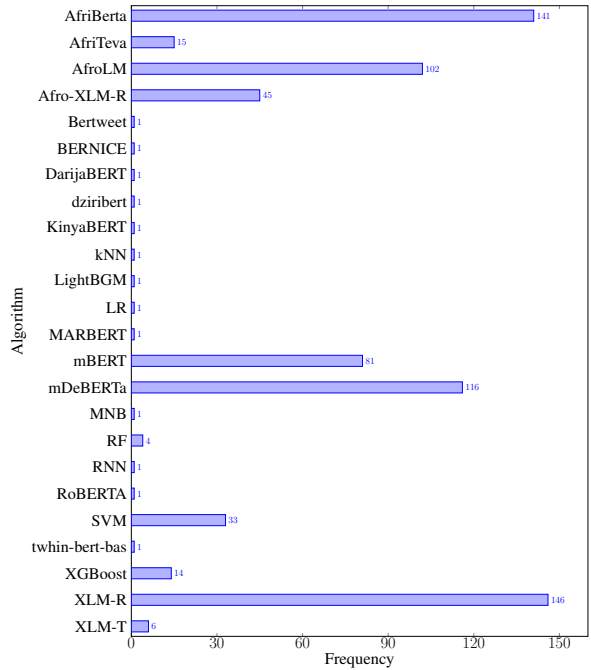
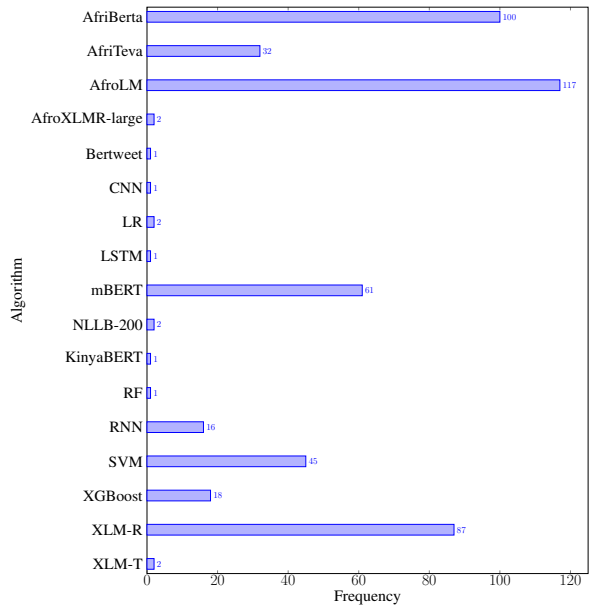
\begin{figure}[htb]
    \centering
    \subfloat[Classification]{
        \resizebox{\columnwidth}{!}{%
            \begin{tikzpicture}
            \begin{axis}[xbar,
                width=22cm,
                height=30cm,
                xmin=0,
                xmax=160,        
                ylabel={Algorithm},
                ylabel near ticks,
                xlabel={Frequency},
                xlabel near ticks,
                ytick=data,
                bar width=0.5cm,
                enlarge y limits={abs=0.45cm},
                xtick={0,30,...,150},scaled x ticks=false,
                symbolic y coords={AfriBerta, AfriTeva, AfroLM, Afro-XLM-R, Bertweet, BERNICE, DarijaBERT, dziribert, KinyaBERT, kNN, LightBGM, LR, MARBERT, mBERT, mDeBERTa, MNB, RF, RNN, RoBERTA, SVM, twhin-bert-bas, XGBoost, XLM-R, XLM-T},
                major y tick style = {opacity=0},
                minor y tick num = 1,
                minor tick length=2ex,
                ylabel style={font=\fontsize{21}{24}\selectfont},
                xlabel style={font=\fontsize{21}{26}\selectfont},
                yticklabel style = {font=\huge,xshift=0.5ex},
                xticklabel style = {font=\huge,yshift=0.5ex},
                nodes near coords,  
                every node near coord/.append style={font=\large},
                y dir=reverse
            ]
            \addplot table[x index=1,y index=0] \algorithmClassStat;
        \end{axis}
        \end{tikzpicture}
        }
    }\\
    \subfloat[Pre-processing]{
        \resizebox{\columnwidth}{!}{%
            \begin{tikzpicture}
            \begin{axis}[xbar,
                width=22cm,
                height=28cm,
                xmin=0,
                xmax=125,        
                ylabel={Algorithm},
                ylabel near ticks,
                xlabel={Frequency},
                xlabel near ticks,
                ytick=data,
                bar width=0.5cm,
                enlarge y limits={abs=0.45cm},
                xtick={0,20,...,120},scaled x ticks=false,
                symbolic y coords={AfriBerta, AfriTeva, AfroLM, AfroXLMR-large, Bertweet, CNN, LR, LSTM, mBERT, NLLB-200, KinyaBERT, RF, RNN, SVM, XGBoost, XLM-R, XLM-T},
                major y tick style = {opacity=0},
                minor y tick num = 1,
                minor tick length=2ex,
                ylabel style={font=\fontsize{21}{24}\selectfont},
                xlabel style={font=\fontsize{21}{26}\selectfont},
                yticklabel style = {font=\huge,xshift=0.5ex},
                xticklabel style = {font=\huge,yshift=0.5ex},
                nodes near coords,  
                every node near coord/.append style={font=\large},
                y dir=reverse
            ]
            \addplot table[x index=1,y index=0] \algorithmPreProcStat;
        \end{axis}
        \end{tikzpicture}
        }
    }
    \caption{Algorithms used for pre-processing and final tweet classification.}
    \label{fig:algorithm_stats}
\end{figure}

%% file: tools-used.tex
\pgfplotstableread[row sep=\\,col sep=&]{
    Tools & Frequency\\
    hugginface & 36\\
    pytorch & 29\\
    pandas & 27\\
    scikit-learn & 23\\
    nltk & 8\\
    tensorflow & 6\\
    keras & 5\\
    pytorch-lightning & 3\\
    glove & 2\\
    emoji & 2\\
}\toolsUsed

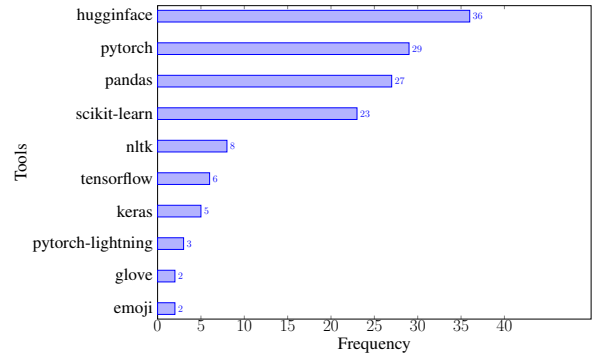
\begin{figure}[htb]
    \centering
    \resizebox{\columnwidth}{!}{%
            \begin{tikzpicture}
            \begin{axis}[xbar,
                width=20cm,
                height=15cm,
                xmin=0,
                xmax=50,        
                ylabel={Tools},
                ylabel near ticks,
                xlabel={Frequency},
                xlabel near ticks,
                ytick=data,
                bar width=0.5cm,
                enlarge y limits={abs=0.45cm},
                xtick={0,5,...,40},scaled x ticks=false,
                symbolic y coords={hugginface, pytorch, pandas, scikit-learn, nltk, tensorflow, keras, pytorch-lightning, glove, emoji},
                major y tick style = {opacity=0},
                ylabel style={font=\fontsize{21}{24}\selectfont},
                xlabel style={font=\fontsize{21}{26}\selectfont},
                yticklabel style = {font=\huge,xshift=0.5ex},
                xticklabel style = {font=\huge,yshift=0.5ex},
                nodes near coords,  
                every node near coord/.append style={font=\large},
                y dir=reverse
            ]
            \addplot table[x index=1,y index=0] \toolsUsed;
        \end{axis}
        \end{tikzpicture}
        }
    \caption{Tools Used}
    \label{fig:tools_used}
\end{figure}

%% file: affiliation.tex
\begin{table*}[h]
    \centering
    \resizebox{\textwidth}{!}{
    \begin{tabular}[t]{|l|l|l|p{11cm}|}
    \hline
        \textbf{Tasks} & \textbf{Team} & \textbf{System paper} & \textbf{Affiliation} \\
    \hline
        A,B,C & ABCD Team & \cite{dang-EtAl:2023:SemEval} & University of Information Technology - Ho Chi Minh City, Vietnam; National University, Ho Chi Minh City, Vietnam; Nong Lam University - Ho Chi Minh City, Vietnam\\\hline
        B,C & BERT 4EVER & --- & Guangdong University of Technology; Guangdong University of Foreign Studies\\\hline
        A,C & Bhattacharya\_Lab & \cite{hughes-EtAl:2023:SemEval} & Auburn University at Montgomery; Florida Polytechnic University\\\hline
        A,B,C & DN & \cite{homskiy-maloyan:2023:SemEval} & none\\\hline
        A,B & DuluthNLP & \cite{akrah-pedersen:2023:SemEval} & University of Minnesota, Duluth, USA\\\hline
        A,B,C & FIT BUT & \cite{aparovich-EtAl:2023:SemEval} & Brno Technical University, Czech Republic\\\hline
        A & Foul & \cite{belbachir:2023:SemEval} & IPSA - Ecole d’ingenieurs A\'eronautique et Spatiale Paris\\\hline
        A,B,C & FUOYENLP & --- & Federal University Oye-Ekiti Ekiti, Nigeria\\\hline
        A,B,C & GMNLP & \cite{alam-EtAl:2023:SemEval} & George Mason University\\\hline
        A,B,C & GunadarmaxBRIN & \cite{arlim-EtAl:2023:SemEval} & Gunadarma University, Indonesia; National Research and Innovation Agency, Indonesia\\\hline
        A,B,C & HausaNLP & \cite{abdullahi-EtAl:2023:SemEval} & Kaduna State University; University of Bucharest; University of Abuja; Nile University; Universiti Teknologi PETRONAS; Bayero University Kano; Shehu Shagari College of Education; Ahmadu Bello University; Masakhane; HausaNLP\\\hline
        A,B & Howard University CS & \cite{aryal-prioleau:2023:SemEval} & Howard University\\\hline
        B & iREL & --- & IIIT-H\\\hline
        B & ISCL\_WINTER & \cite{hancharova-wang-kumar:2023:SemEval} & Eberhard Karls Universität Tübingen\\\hline
        A,B & JCT & \cite{keinan-hacohenkerner:2023:SemEval} & Jerusalem College of Technology\\\hline
        A & KINLP & \cite{nzeyimana:2023:SemEval} & University of Massachusetts Amherst\\\hline
        A,B & king001 & --- & PingAnLifeInsurance\\\hline
        A,C & MaChAmp & \cite{vandergoot:2023:SemEval} & none \\\hline
        A,B,C & Masakhane-Afrisenti & \cite{azime-EtAl:2023:SemEval} & Saarland Universty; Luleå University of Technology; Instituto Politécnico Nacional; Masakhane; Montclair State University; University of Ibadan; Iowa State University\\\hline
        A,B,C & mitchelldehaven & --- & Information Sciences Institute (ISI)\\\hline
        A,B,C & NLNDE & \cite{wang-EtAl:2023:SemEval2} & Bosch Center for Artificial Intelligence, Renningen, Germany; Center for Information and Language Processing (CIS), LMU Munich, Germany\\\hline
        A,B,C & NLP-LISAC & \cite{benlahbib-boumhidi:2023:SemEval} & LISAC Faculty of Sciences Dhar EL Mehraz (F.S.D.M), Fez, Morocco\\\hline
        C & PA2023 & --- & PingAn\\\hline
        A,B,C & PA14 & --- & PA Sales Support Development  Co., Ltd.\\\hline
        A,B,C & PingAnLifeInsurance & \cite{jin-EtAl:2023:SemEval} & Ping An Life Insurance Company of China, Ltd.\\\hline
        B,C & saroyehun & --- & University of Konstanz\\\hline
        A & Seals\_Lab & \cite{raychawdhary-EtAl:2023:SemEval} & Auburn University\\\hline
        A,B,C & Sefamerve & \cite{delil-kuyumcu:2023:SemEval} & Sefamerve R\&D Center\\\hline
        C & Snarci & --- & University of Cagliari\\\hline
        A & TechSSN & \cite{sivanaiah-EtAl:2023:SemEval} & Sri Sivasubramaniya Nadar College of Engineering, Chennai, India\\\hline
        A,B,C & tmn & --- & University of Tyumen, Russia\\\hline
        A & Trinity & \cite{rathi-EtAl:2023:SemEval} & Pune Institute Of Computer Technology\\\hline
        A,C & UBC-DLNLP & \cite{bhatia-EtAl:2023:SemEval} & The University of British Columbia; Mohamed Bin Zayed University of Artificial Intelligence\\\hline
        A,B,C & UCAS-IIE-NLP & \cite{hu-EtAl:2023:SemEval} & Chinese Academy of Sciences; University of Chinese Academy of Sciences\\\hline
        A,B & UIO & \cite{rnningstad:2023:SemEval} & University of Oslo\\\hline
        A,B,C & UM6P & \cite{elmahdaouy-EtAl:2023:SemEval} & Mohammed VI Polytechnic University, Morocco; University of Luxembourg, Luxembourg\\\hline
        A,B & UMUTeam & \cite{garcadaz-EtAl:2023:SemEval1} & University of Murcia; University of Castilla-La Mancha\\\hline
        A & Uppsala University & \cite{kniele-beloucif:2023:SemEval} & Uppsala University\\\hline
        A,B & Witcherses & \cite{gokani-srivatsa-mamidi:2023:SemEval} & Language Technologies Research Center (LTRC); Kohli Center on Intelligent Systems; International Institute of Information Technology, Hyderabad\\\hline
        A,B,C & ymf924 & --- & The Ohio State University\\
    \hline
    \end{tabular}
    }
    \caption{Participating teams, their affiliations and the tasks they participated in.}
    \label{tab:affiliation}
\end{table*}